\documentclass[11pt]{article}

\usepackage{graphicx}
\usepackage{amssymb,amsmath,amsfonts}
\usepackage{epsf}
\usepackage{wrapfig}
\usepackage[normalem]{ulem}
\usepackage[mathscr]{euscript}
\usepackage{mathtools}
\usepackage{ifmtarg}
\usepackage[utf8]{inputenc}
\usepackage{fancyvrb}

\newcommand{\R}{\mathbb{R}}

\newcommand{\E}{\mathbb{E}}

\newcommand{\pr}{\mathbb{P}}

\usepackage{color}

\DefineVerbatimEnvironment{Latex}{Verbatim}{numbers=left,numbersep=2mm}

\makeatletter

    \newcommand\contFrac{\@ifstar{\@contFracStar}{\@contFracNoStar}}

    \def\singleContFrac#1#2{%
        \begin{array}{@{}c@{}}%
            \multicolumn{1}{c|}{#1}%
            \\%
            \hline%
            \multicolumn{1}{|c}{#2}%
        \end{array}%
    }

    \def\@contFracNoStar#1{%
        \mathchoice{
            \@contFracNoStarDisplay@#1//\@nil%
        }{
            \@contFracNoStarInline@#1//\@nil%
        }{
            \@contFracNoStarInline@#1//\@nil%
        }{
            \@contFracNoStarInline@#1//\@nil%
        }%
    }

    \def\@contFracNoStarDisplay@#1//#2\@nil{%
        \@ifmtarg{#2}{%
            #1%
        }{%
            #1+\cfrac{1}{\@contFracNoStarDisplay@#2\@nil}%
        }%
    }

        \def\@contFracNoStarInline@#1//#2\@nil{%
            \@ifmtarg{#2}{%
                #1%
            }{%
                #1 \@@contFracNoStarInline@@#2\@nil%
            }%
        }
        \def\@@contFracNoStarInline@@#1//#2\@nil{%
            \@ifmtarg{#2}{%
                + \singleContFrac{1}{#1}%
            }{%
                + \singleContFrac{1}{#1} \@@contFracNoStarInline@@#2\@nil%
            }%
        }

    \def\@contFracStar#1{%
        \mathchoice{
            \@contFracStarDisplay@#1////\@nil%
        }{
            \@contFracStarInline@#1//\@nil%
        }{
            \@contFracStarInline@#1//\@nil%
        }{
            \@contFracStarInline@#1//\@nil%
        }%
    }

    \def\@contFracStarDisplay@#1//#2//#3\@nil{%
        \@ifmtarg{#2}{%
            #1%
        }{%
            #1 + \cfrac{#2}{\@contFracStarDisplay@#3\@nil}%
        }%
    }

        \def\@contFracStarInline@#1//#2\@nil{%
            \@ifmtarg{#2}{%
                #1%
            }{%
                #1 \@@contFracStarInline@@#2\@nil%
            }%
        }
        \def\@@contFracStarInline@@#1//#2//#3\@nil{%
            \@ifmtarg{#3}{%
                - \singleContFrac{#1}{#2}%
            }{%
                - \singleContFrac{#1}{#2} \@@contFracStarInline@@#3\@nil%
            }%
        }
\makeatother

\usepackage{color}

\usepackage{ulem}

\usepackage{blkarray}

\usepackage{lscape}

\usepackage{cancel}

\usepackage{setspace}

\usepackage{caption}
\usepackage{subcaption}

\usepackage{csquotes}
\usepackage{fancyhdr}
\usepackage{lastpage}
\usepackage{enumerate}

\usepackage{color,soul}

\usepackage{mdframed}

\usepackage[version=4]{mhchem}

\usepackage{bbm}

\DeclareMathOperator*{\argmax}{arg\,max}
\DeclareMathOperator*{\argmin}{arg\,min}

\usepackage{enumitem}

\usepackage{authblk}

\usepackage[ruled,vlined]{algorithm2e}

\usepackage{mathtools}
\mathtoolsset{showonlyrefs}

\allowdisplaybreaks

\usepackage{booktabs}       

\usepackage{multirow}

\usepackage{hhline}

\title{Open-Set Recognition of Breast Cancer Treatments}

\author[1]{Alexander Cao\thanks{Corresponding author: Alexander Cao, Technological Institute C138, 2145 Sheridan Rd, Evanston, IL}}
\author[1]{Diego Klabjan}
\author[2]{Yuan Luo}
\affil[1]{Department of Industrial Engineering and Management Sciences, Northwestern University, Evanston, IL}
\affil[2]{Department of Preventive Medicine, Northwestern University, Chicago, IL}
\affil[ ]{a-cao@u.northwestern.edu, \{d-klabjan, yuan.luo\}@northwestern.edu}

\begin{document}
\maketitle
\thispagestyle{fancy}

\section*{Keywords}
Breast cancer treatments, open-set recognition, deep neural networks, autoencoder 

\section*{Abstract}
Open-set recognition generalizes a classification task by classifying test samples as one of the known classes from training or ``unknown.'' As novel cancer drug cocktails with improved treatment are continually discovered, predicting cancer treatments can naturally be formulated in terms of an open-set recognition problem. Drawbacks, due to modeling unknown samples during training, arise from straightforward implementations of prior work in healthcare open-set learning. Accordingly, we reframe the problem methodology and apply a recent existing Gaussian mixture variational autoencoder model, which achieves state-of-the-art results for image datasets, to breast cancer patient data. Not only do we obtain more accurate and robust classification results, with a 24.5\% average F1 increase compared to a recent method, but we also reexamine open-set recognition in terms of deployability to a clinical setting. 

\section{Introduction}
Most prior work in classification is closed-set; meaning the classes are assumed to be the same for both training and testing. Only relatively recently have classifiers designed for open-set evaluation, where unknown classes appear only in testing, gained attention as a real-world necessity. In particular, open-set image recognition arises from increasingly automated computer vision systems such as those in self-driving cars. It would certainly be impossible to include every object class that could possibly be seen while driving in model training \cite{survey}. Given the inherent dynamism of healthcare, one can argue a greater need for open-set classifiers. Some diseases are too rare to include sufficient samples in training \cite{oct_images}. There is also a consistent cycle of identifying novel diseases and developing treatments for them (recently COVID-19, for instance). Both of these circumstances necessitate generalizing medical classification tasks to open-set recognition. In this paper we study an example from a more prevalent circumstance: the discovery of new drug combinations for already existing diseases.

Personalized medicine through quantitative, phenotypic profiling shows promise in medical care by guiding drug combination strategies \cite{ashtma-phenotype-personalized-medicine}. In cancer treatments, these drug combinations are becoming the standard of care and many drug combination therapies have been approved or are under clinical trials \cite{need-cancer-treatment-combinations}. The landscape of cancer drug combinations, or ``cocktails,'' evolves with discoveries of novel cocktails with improved treatment and lessening side effects. Although some guidelines exist for certain cancer types, individual patients’ responses to various drug combinations are still not well understood. For instance, which drug combinations would likely benefit a specific patient the most is still a critical, open question. In this vein, we formulate predicting cancer cocktail treatments as an open-set classification problem. Our goal is to classify patients by cocktail treatments based on medical and demographic features. In addition, however, sufficiently unique patients unlike those historically associated with known cocktails (i.e., cocktails in the training set) should be classified as ``novel.'' This ``novel" class is an indication that different or new cocktails may be more suitable for those patients’ treatments. To our knowledge, this is the first application of open-set recognition to cancer treatment predictions.

In this paper, we focus on the open-set learning variant of training (and validating) on only the $C$ known classes for $(C+1)$-class classification during inference. The $(C+1)$-th class aggregates all novel test samples not belonging to the known classes. To reflect a real-world scenario, we do not have samples from ``novel" cocktails during the training and validation phases. For this pilot study, we focus on predicting if a patient will benefit from a novel cocktail treatment versus known cocktails. Previous healthcare open-set studies rely on the use of fabricated or auxiliary data and standard softmax classifiers \cite{oct_images, diagnoses}. Bypassing this data necessity and instead exploiting reconstruction, we adapt the existing Gaussian Mixture Variational Autoencoder (GMVAE) model \cite{my-gmvae}, which achieves state-of-the-art results in open-set image recognition, to our open-set cancer treatment recognition task. GMVAE is a deep neural network, autoencoder-based model in which the bottleneck latent layer simultaneously performs class-based clustering and learns reconstruction. In this way, the patient data is embedded in a lower-dimensional representation that discriminates between known cocktail classes and, unlike standard approaches, simultaneously captures interdependent patient information. This dual nature of classification and reconstruction leads to a more flexible and amenable latent representation. With the model's embedding in hand,  \cite{my-gmvae} applies an ``uncertainty'' threshold based on distances to class centroids to more accurately and robustly distinguish between the known and ``novel'' cocktail classes. We apply these methods to breast cancer patients' electronic health records (EHRs) from Northwestern Memorial Hospital. In doing so, this study achieves a step towards implementing a system to help physicians identify cancer patients who may benefit from a novel drug cocktail in a real-time clinical setting. 

Our paper is organized as follows. In \S \ref{section:related-work}, we compare related work along with a comprehensive summary of the benchmark model \cite{inter-intra}. Following in \S \ref{section:methodology}, we first provide background on the GMVAE model coupled with the ``uncertainty'' threshold \cite{my-gmvae}. In particular, we emphasize the intuition behind dual reconstruction-classification learning and ``uncertainty'' for open-set recognition. Next, we present the complete experimental design from data feature engineering to model evaluation. Subsequently in \S \ref{section:results}, we conduct open-set recognition experiments on our breast cancer patient dataset. 

From these experimental results, we stress two findings, which are the main contributions. First, GMVAE outperforms a state-of-the-art, solely classification-based, deep open-set classifier both in terms of accuracy and robustness to an increasing number of unknown cocktails. Second, relevant prior methods \cite{oct_images, diagnoses, inter-intra} bypass selecting a single optimal threshold for rejecting unknowns by reporting AUC or ROC metrics or simply assuming a binary known-unknown false positive rate. However, all of these are uninformative for actual model deployment where a single threshold would be used for decisions. In contrast, GMVAE combined with ``uncertainty'' showcases an intuitive heuristic for selecting a single, optimal threshold. This process fits a threshold based on the known validation set classification accuracies and is further explained in \S \ref{section:results}. We emphasize this is a more practical model evaluation comparison. Summary ROC metrics can be useful in comparing different models in a holistic sense. However, in terms of real-world model usage in a clinical setting, it is more apt to compare actual decision accuracies which are only apparent after choosing a threshold. Finally in \S \ref{section:limitations-future_work} and \S \ref{section:conclusion}, we end with a discussion on limitations and future work, and conclude. 

\section{Related work} \label{section:related-work}
For literature placement, it is important to note that open-set recognition reduces to outlier detection in the case where the number of known classes $C=1$ (viewed as a ``normal'' class). Outlier detection, or the related novelty or anomaly detection, is a longer studied topic \cite{survey, anomaly, outlier-exposure}. Such methods are utilized in healthcare to detect outliers in breast cancer survivability predictions \cite{outlier-breast_cancer-survival} and anomalous activity in EHRs \cite{outlier-epr}. Outlier detection, however, does not generally extend to differentiating between multiple known classes hence open-set recognition. For instance, in our breast cancer patient experiments there are three and four known cocktail classes. 

There is an immense body of existing work concerning traditional closed-set classification. Open-set recognition, on the other hand, is only relatively recently receiving more consideration. Earlier examples of $(C+1)$-class classification employ SVMs \cite{TowardsOpenSetRecognition, ProbabilityofInclusion} or sparse representation \cite{sparse}. Open-set recognition in conjunction with deep neural networks is the current trend \cite{oct_images, TowardsOpenSetDeepNetworks, crosr, c2ae}. However, these methods are almost exclusively designed solely for image recognition; network architecture reliance on image patching, channel activation, spatial pooling, feature map modulation, and pixel reconstruction inhibit usability for non-image-based tasks (such as ours). 

While image classification benefits from a well-rounded surge in open-set recognition, applications to general healthcare data are wanting. Specifically in \cite{oct_images}, eye diseases are open-set classified using optical coherence tomography (OCT) images but the method is contingent on a patchGAN-derived model \cite{patchGAN} to generate synthetic, ``boundary'' images that are deliberately difficult to classify with a pretrained, closed-set softmax classifier. These manufactured outliers are then added to the original dataset and used to train a standard $(C+1)$-class classifier. The multi-phase training, known complications of training GANs, and assumed image-based data limit the generalization of this work to other healthcare applications. Furthermore, the authors in \cite{oct_images} visualize the generated ``unknown'' class images to verify they are ``different yet reasonable,'' it is not clear how to apply this criterion to patient demographics or abnormal lab tests that comprise our data. Relatedly, \cite{diagnoses} proposes framing medical diagnosis classification in terms of open-set recognition. Their method treats samples from less common conditions as a proxy for the unknown classes and instead maximizes their cross-entropy during classic softmax training. During inference, a simple threshold is applied to closed-set, softmax probabilities to reject unknown samples. A shortcoming of this method is the restricting assumption that one's dataset can afford such a large enough and representative residual subset. Indeed in \cite{diagnoses}, the authors have a known training set of 160 diagnosis classes and a counterpart set of another 160 diagnoses (each with at least 10\% of training diagnosis' samples) to model the unknown classes. For our breast cancer patient dataset, there are orders of magnitude differences in the number of samples per cocktail, as well as a drug approval timeline. Both reasons render our residual samples inadequate and possibly time-inconsistent for such a procedure.

In contrast, GMVAE naturally serves non-image data and entirely circumvents the need for artificial ``unknown'' or ``novel" samples. Accordingly, for a compatible comparison, we benchmark GMVAE against the so-called ii-loss and outlier score method of \cite{inter-intra}. This specific benchmark is also fitting because it (i) attains state-of-the-art open-set recognition accuracies on two non-image-based datasets, and (ii) makes use of similar latent space distance-based thresholding to reject ``novel'' samples. It is worth noting that \cite{inter-intra} demonstrates that naive thresholding on closed-set, softmax classifiers can lead to significantly poorer open-set recognition. The ii-loss is still wholly classification-based and by contrast GMVAE has the advantage of dual classification-reconstruction learning.

For completeness, we now summarize the ii-loss and outlier score. The authors in \cite{inter-intra} argue that open-set recognition is most amenable in a data embedding that clusters samples from the same known class tightly together (low intra-spread) but pushes samples from different known classes far apart from each other (high inter-spread). To directly produce such a neural network mapping $z$ of the data $x$, they minimize the following loss function:
\begin{align}
\text{ii-loss} = \left(\frac{1}{N} \sum_{i=1}^C \sum_{j=1}^{|C_i|} ||z(x_j) - \mu_i||_2^2 \right)- \left(\min_{i \neq j} ||\mu_i - \mu_j||_2^2\right) \label{eq:ii-loss}
\end{align}
where $N$ is the total number of samples, $|C_i|$ is the number of samples in class $i=1,...,C$ and 
\begin{align}
\mu_i = \frac{1}{|C_i|} \sum_{j=1}^{|C_i|} z(x_j)
\end{align}
is the centroid of each class $i$. The first term in \eqref{eq:ii-loss} measures intra-spread and so aims to minimize the distance between each latent $z$ and its own centroid. The second term in \eqref{eq:ii-loss} quantifies inter-spread and seeks to maximize the minimum distance between class centroids. Batch normalization layers prevent this term from diverging to infinity. The neural network projection $z(x)$ has no set architecture and can be composed of any architectural designs. 

With a trained latent representation $z(x)$ in hand, the outlier score, or squared distance to the nearest centroid, is given by 
\begin{align}
\text{outlier score} \left(\widehat{x} \right) = \min_i || \mu_i - z\left(\widehat{x} \right) ||_2^2
\end{align}
for a test sample $\widehat{x}$. Consequently, distances to centroids also naturally emit a softmax posterior class probability 
\begin{align}
\pr  \left(y = i | \widehat{x}\right) = \frac{\exp \left\{ -||\mu_i -  z\left(\widehat{x} \right) ||_2^2 \right\}}{\sum_{j=1}^C \exp \left\{ -||\mu_j -  z\left(\widehat{x} \right)||_2^2 \right\}}~.
\end{align}
Finally, thresholding on the outlier score, the open-set prediction is
\begin{align}
\widehat{y} =  \begin{cases}
\argmax_i \pr  \left(y = i | \widehat{x}\right)  \quad & \text{if $\text{outlier score} \left(\widehat{x} \right) \leq \tau$} \\
C+1 & \text{otherwise}.
\end{cases}
\end{align}
We argue that a drawback of this entire procedure is the unsystematic, ad-hoc method of selecting the threshold $\tau$. It is assumed that some percentage, a so-called contamination ratio $\alpha$, of the training set are outliers. Correspondingly, the threshold $\tau$ is set to the $1-\alpha$ percentile of all training outlier scores. In experiments, \cite{inter-intra} finds that a 1\% contamination ratio is broadly suitable. While this is certainly easily understood for the user, it lacks any guidance from the embedding clustering and simply follows from the early presumption. In \S \ref{section:results}, we illustrate a more deliberate selection for GMVAE's ``uncertainty'' threshold $\tau$.

\section{Methodology} \label{section:methodology}
In this section we outline the complete methodology used for the open-set recognition of breast cancer treatments experiments. First, we describe the GMVAE model; second, each step of the dataset construction is detailed. Finally, we summarize model training and evaluation procedures. 

\subsection{GMVAE and ``uncertainty'' algorithm}
While the detailed derivation and technical mathematics of GMVAE and the ``uncertainty'' algorithm can be found in \cite{my-gmvae}, we also briefly overview the model in the Appendix. Additionally, in \cite{my-gmvae}, the authors spend a great time justifying the need for a Gaussian mixture embedding per class for images as well as a procedure for identifying the number of components. However, for our breast cancer dataset, initial assessments indicate that we cannot discern enough patient encounter heterogeneity to warrant multiple components per drug cocktail treatment class. Accordingly, we utilize $K=1$ (one cluster per class) for all experiments and simplify GMVAE to a single Gaussian prior for each class. 

\begin{figure}[h!]
\centerline{
  \includegraphics[width=0.95\linewidth]{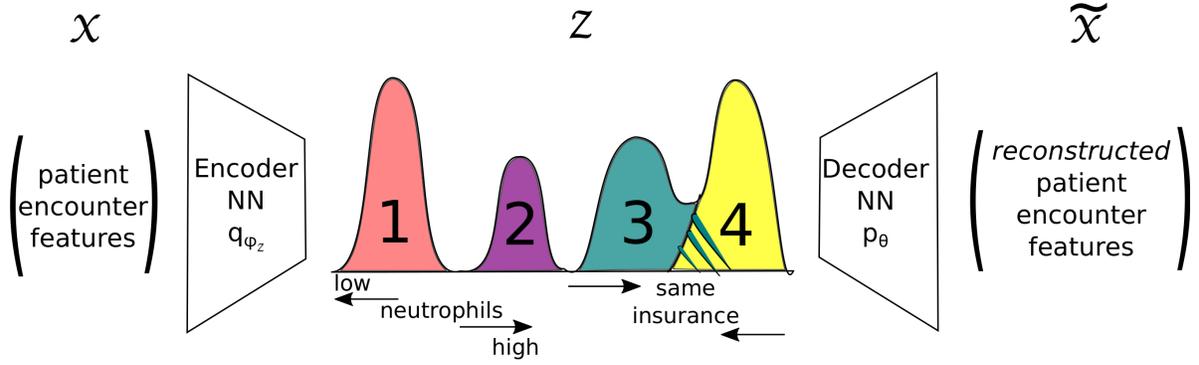} 
}
\caption{One-dimensional depiction of GMVAE's class-based and reconstruction bottleneck for our breast cancer patients dataset. Neural network $q_{\phi_z}$ encodes patient encounter features $x$ into learned embedding $z$. The class numbers (and colors) correspond to drug cocktail treatments prescribed for patients' encounters. Latent variable $z$ is then used to reconstruct the original data $\widetilde{x} \sim x$ via network $p_\theta$.}
\label{fig:explain-gmvae}
\end{figure}

The bulk of this section focuses on an intuitive understanding of GMVAE and ``uncertainty,'' as it relates to the open-set recognition of breast cancer cocktail treatments. While GMVAE's structure is more intricate than a standard VAE, its essence can still be understood as the encoder-decoder composition. The principal difference with unsupervised VAEs is that the latent, bottleneck layer cooperatively performs class-based clustering (clinically can be thought of as endophenotyping) and learns reconstruction. We illustrate this duality in Figure \ref{fig:explain-gmvae}.

Patient encounter features $x$ are projected to latent space $z$ of significantly fewer dimensions hence the bottleneck, with neural network $q_{\phi_z}$. In $z$-space, GMVAE ELBO's latent covering term clusters drug cocktail treatment classes together as depicted with class numbers and colors. However,  these class clusters are translated and scaled by the reconstruction term (see Appendix) which promotes patient encounters with similar features to be closer and vice-versa. For example, drug cocktails 1 and 2 may be separated based on neutrophil levels and this characteristic further discriminates these classes. Conversely, a subset of patients of drug cocktails 3 and 4 may share the same insurance, forcing the class clusters to overlap (shown as the yellow-green striped region). While this overlap may be seen as counterproductive, we believe one should not weight or select features to maximize class separation. The reason being that one does not know apriori which features will best separate the ``novel" samples. Therefore, an embedding $z$ which most accurately represents data features will naturally separate those distinguishing ``novel'' samples. Finally, samples' representations $z$ are used to reconstruct the features $\widetilde{x}$ via network $p_\theta$. Of course this cooperative, multi-task learning occurs across the entire multi-dimensional $z$-space. It is important to note here that because of GMVAE's tendency to overlap classes based on reconstruction, the ii-loss exhibits better discrimination among the known classes. However, this closed-set weakness becomes an open-set strength as this behavior shrinks high-risk open-space between the known clusters.

\begin{figure}[h!]
\centerline{
  \includegraphics[width=0.5\linewidth]{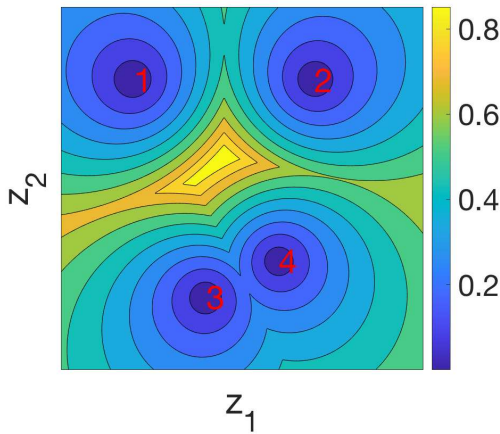} 
}
\caption{Two-dimensional heat-map visualization of the ``uncertainty'' $U$ with four known centroids labeled with red numbers.}
\label{fig:uncertainty-example}
\end{figure}

While GMVAE's encoder produces a latent space Gaussian (outputs mean and diagonal covariance), the mean $\mu(x; \phi_z)$ is instinctively designated as the effective $z$-space mapping. Similar to the outlier score, \cite{my-gmvae} then applies a distance measure to carve the ``novel'' decision boundaries around each known centroid. The ``uncertainty'' threshold quantity is defined as the ratio between the distance to the nearest centroid and the average distance to all other centroids. For our $K=1$ case with test sample $\widehat{x}$, let us denote $\overline{z}_c$ as each known class's training latent centroid and $c^* = \argmin_c || \mu \left(  \widehat{x}; \phi_z \right) - \overline{z}_c||_2$. Then ``uncertainty'' $U$ is mathematically expressed as 
\begin{align}
U = \frac{|| \mu \left(  \widehat{x}; \phi_z \right) - \overline{z}_{c^*}||_2}{\frac{1}{C-1} \sum_{c\neq c^*}|| \mu \left(  \widehat{x}; \phi_z \right) - \overline{z}_c||_2}
\end{align} 
with the corresponding classification rule
\begin{align}
\widehat{y} =  \begin{cases}
c^*  \quad & \text{if $U \leq \tau$} \\
C+1 & \text{otherwise}.
\end{cases} 
\end{align}
The key differences are that this threshold captures orientation with respect to known centroids (unlike the rotationally symmetric outlier score) and is scale invariant. We visualize these attributes in Figure \ref{fig:uncertainty-example}.

For non-trivial open-set recognition, we may assume the ``unknown'' or ``novel" samples are comparable to the known classes. As such, there exists a large risk of incorrectly predicting a ``novel'' sample as one of the known classes. Thresholding upon $U$ seeks to minimize this risk by penalizing the open space between known centroids more heavily, as perceived in Figure \ref{fig:uncertainty-example}. If $U=0$ then the test sample's latent embedding is exactly one of the known centroids with no doubt of its classification. However, if $U=1$ then the test sample's embedding is equidistant to all known centroids and is unclassifiable among the known classes. In addition, $U$ approaches 1 if the test sample's embedding is sufficiently far from all known centroids. Finally, the metric $U$ is designed to be standardized between 0 and 1, making it more universally applicable as opposed to the outlier score's raw distance.

\subsection{Dataset construction}
The dataset consists of breast cancer patient records at Northwestern Memorial Hospital spanning from 2000 to 2015. We consider patient encounters as independent samples. While this removes the longitudinal aspect of the data, it allows for a more direct application of existing, timeless open-set recognition methods and also matches the cancer treatment narrative because physicians can adjust drug cocktail treatments as patients respond differently or side effects flare. 

The samples' classes are specific drug cocktails and are assembled by simply aggregating the prescribed medications for each patient encounter. We only include drugs principally related to treating cancer (as listed by the National Cancer Institute). For the purposes of our breast cancer-related task, drugs like Acetaminophen are extraneous and therefore excluded. In addition, we ultimately only take those cocktails with more than 1,000 encounters to maintain reasonably-sized classes. Table \ref{tab:cocktails} below summarizes the drug cocktail classes of our dataset. Each cocktail's FDA approval year is set to its latest component drug's FDA approval year. Perhaps not so surprising is that most encounter-level cocktails are composed of only a single drug. Even so, we refer to these as cocktails for convenience.

\begin{table}[h!]
\centering
\caption{Breast cancer drug cocktail classes ordered by FDA approval year.}
\label{tab:cocktails}
\begin{tabular}{ cccc }   
Cocktail & Drugs  & Number of   & FDA \\  
number &   & encounters  & approval year \\
\begin{tabular}[t]{ c } 
\hline
1 \\
2 \\ 
3 \\
4 \\
5 \\
6 \\
7 
\end{tabular} &  
\begin{tabular}[t]{ l} 
\hline
Dexamethaosne, Ondansetron \\
Ondansetron \\
Paclitaxel \\
Trastuzumab \\
Pegfilgrastim \\
Cyclophosphamide, Doxorubicin, Palonosetron \\
Denosumab 
\end{tabular} & 
\begin{tabular}[t]{ c} 
\hline
2,714 \\
1,868 \\
3,553 \\
5,454  \\
1,977 \\ 
2,416 \\ 
2,616 
\end{tabular} & 
\begin{tabular}[t]{ c} 
\hline
1991 \\
1991 \\
1992 \\
1998  \\
2002 \\ 
2003 \\ 
2010 
\end{tabular} \\
\end{tabular}
\end{table}

Phenotypic feature engineering for the patient encounters is relatively straightforward with few transformations. We enumerate and categorize demographic and physical characteristic features in Table \ref{tab:demographics}, diagnoses (ICD-9 codes) in Table \ref{tab:icd9}, and lab features in Table \ref{tab:labs}. All demographics except ``age at encounter'' are the same for a single patient across encounters. Physical characteristics, having little variance, are averaged across encounters for each patient. In addition, we cutoff ICD-9 codes with less than 1,000 total encounters so that diagnoses are relevant to the dataset as a whole.

\begin{table}[h!]
\centering
\caption{Distributions of demographic and physical characteristic features of breast cancer patient encounters. \textbf{Bolded} features are consistent per patient. Categorical demographics enumerated with indents.}
\label{tab:demographics}
\begin{tabular}{ cc }   
Demographics/ & Count (percent)/ \\
Physical characteristics & Median (IQR) \\
\begin{tabular}[t]{ l } 
\hline
\textbf{Race} \\
\hspace{5pt} Native American \\
\hspace{5pt} Asian \\
\hspace{5pt} Black \\
\hspace{5pt} Hispanic \\
\hspace{5pt} White \\
\hspace{5pt} Unknown \\
\textbf{Ethnicity} \\ 
\hspace{5pt} Hispanic or Latino \\
\hspace{5pt} Not Hispanic or Latino \\
\hspace{5pt} Unknown \\
\textbf{Marital status} \\
\hspace{5pt} Divorced \\
\hspace{5pt} Married \\
\hspace{5pt} Separated \\
\hspace{5pt} Significant Other \\
\hspace{5pt} Single \\
\hspace{5pt} Widowed \\
\hspace{5pt} Unknown \\
\textbf{Gender} \\
\hspace{5pt} Female \\
\hspace{5pt} Male \\
\textbf{Insurance} \\ 
\hspace{5pt} Private \\
\hspace{5pt} Medicare \\
\hspace{5pt} Medicaid \\
\hspace{5pt} Unknown \\
Age at encounter \\
 \\
\textbf{BMI} \\
\textbf{Height} \\
\textbf{Weight}
\end{tabular} &  
\begin{tabular}[t]{ l} 
\hline
 \\
44 (0.2\%) \\
1,053 (5.1\%) \\
3,246 (15.8\%) \\
255 (1.2\%) \\
13,111 (63.7\%) \\
2,889 (14.0\%) \\
\\
1,488 (7.2\%) \\
17,650 (85.7 \%) \\
1,460 (7.1\%) \\
\\
1,777 (8.6\%) \\
11,699 (56.8\%) \\
59 (0.3\%) \\
3 (0.01\%) \\
5,645 (27.4\%) \\
959 (4.7\%) \\
456 (2.2\%) \\
\\
20,497 (99.5\%) \\
101 (0.5\%) \\
\\
12,456 (60.5\%) \\
3,626 (17.6\%) \\
1,648 (8.0\%) \\
2,868 (13.9\%) \\
53.24  (45.4 - 62.7) \\
 \\
26.60 (23.3 - 30.9) \\
64.0 (62.2 - 66.0) \\
155.46 (135.9 - 181.0)
\end{tabular} 
\end{tabular}
\end{table}

\begin{table}[h!]
\centering
\caption{Diagnosis (ICD-9 code) features of breast cancer patient encounters grouped by disease type.}
\label{tab:icd9}
\begin{tabular}{ c c }   
Disease group & ICD-9 codes \\  
\begin{tabular}[t]{l} 
\hline
Viral diseases accompanied by exanthem \\
\hline
Malignant neoplasms \\
$\left. \right.$ \\
$\left. \right.$ \\
\hline
Benign neoplasms, carcinoma in situ \\
\hline
Diseases of thyroid and other glands\\
\hline
Nutritional deficiencies, \\
metabolic and immunity disorders \\
\hline
Blood diseases \\
\hline
Mental disorders \\
\hline
Diseases of nervous system and \\
sense organs \\
\hline
Hypertensive, ischemic heart,\\
pulmonary circulation diseases\\
\hline
Other heart diseases \\
\hline
Diseases of circulatory system \\
\hline
Diseases of respiratory system \\
$\left. \right.$ \\
\hline
Diseases of digestive system \\
$\left. \right.$ \\
\hline
Diseases of urinary system \\
\hline
Disorders of breasts \\
\hline
Disorders of female genital tract \\
\hline
Contact dermatitis and other eczema \\
\hline
Arthropathies, dorsopathies, rheumatism \\
$\left. \right.$ \\
$\left. \right.$ \\
\hline
Other disorders of bone and cartilage \\
\hline
Other symptoms \\
$\left. \right.$ \\
$\left. \right.$ \\
$\left. \right.$ \\
\hline
Nonspecific (abnormal) findings in blood, \\
radiological examination \\
\hline
Need for vaccine/inoculation \\
against influenza \\
\hline
Personal/family history of malignant \\
neoplasm, other hazards \\
\hline
Other postprocedural states, \\
conditions influencing health
\end{tabular} & 
\begin{tabular}[t]{ l} 
\hline
53.9, 54.9 \\
\hline
153.9, 162.9, 174.8, 174.9, 182.0, 183.0 193.0, \\
196.3, 196.9, 197.0, 97.7, 198.2, 198.3, 198.4, \\
198.5, 199.1, 202.8 \\
\hline
211.3, 217.0, 218.9,  233.0 \\
\hline
241.0, 241.1, 244.9, 250.0 \\
\hline
266.2, 268.9, 272.0, 272.4, 276.8, 278.0 \\
$\left. \right.$ \\
\hline
280.9, 285.22, 285.9 \\
\hline
300.0, 305.1, 311.0 \\
\hline
327.23, 338.3, 346.9, 354.0, 355.9, 356.9, 362.5, \\
365.9, 366.9, 375.15, 389.9 \\
\hline
401.1, 401.9, 414.0, 415.19 \\
$\left. \right.$ \\
\hline
424.0, 424.1, 425.4, 427.31, 428.0, 429.9 \\
\hline
434.91, 443.9, 453.4, 453.9, 455.6, 457.1 \\
\hline
473.9, 477.9, 486.0, 493.9, 496.0, 511.81, 511.9, \\
518.89 \\
\hline
530.81, 553.3, 562.1, 562.11, 564.0, 564.1, \\
571.8, 573.8, 574.2 \\
\hline
592.0, 593.9, 599.0 \\
\hline
610.8, 611.72, 611.79, 612.1 \\
\hline
620.2, 627.2 \\
\hline
692.9 \\
\hline
714.0, 715.0, 715.9, 716.9, 719.41, 719.45, \\
719.46, 723.1, 724.0, 724.02, 724.2, 724.5, \\
729.1, 729.5 \\
\hline
733.0, 733.9 \\
\hline
780.4, 780.52, 780.57, 780.79, 781.2, 782.1, \\
782.3, 782.62, 784.0, 785.1,  785.6, 786.05, \\
786.09, 786.2, 786.5, 787.01, 787.02, 787.91,\\
 788.3, 788.41, 789.0 \\
\hline
790.29, 793.19, 793.8, 793.89 \\
$\left. \right.$ \\
\hline
V04.81 \\
$\left. \right.$ \\
\hline
V10.3, V15.3, V16.3 \\
$\left. \right.$ \\
\hline
V45.71, V45.89, V49.81 
\end{tabular} \\
\end{tabular}
\end{table}

\begin{table}[h!]
\centering
\caption*{Table 3 continued: Diagnosis (ICD-9 code) features of breast cancer patient encounters grouped by disease type.}
\label{tab:icd9-continued}
\begin{tabular}{ c c }   
Disease group & ICD-9 codes \\  
\begin{tabular}[t]{l} 
\hline
Encounter for other procedure \\
\hline
Special examinations, screenings \\
\hline
Genetic susceptibility to disease \\
\hline
Estrogen receptor positive status \\
\hline
Personal history of antineoplastic \\
chemotherapy
\end{tabular} & 
\begin{tabular}[t]{ l} 
\hline
V58.11, V58.69 \\
\hline
V72.31, V72.83, V72.84, V77.91, V82.81 \\
\hline
V84.01, V84.02 \\
\hline
V86.0 \\
\hline
V87.41 \\
$\left. \right.$ 
\end{tabular} \\
\end{tabular}
\end{table}

\begin{table}[h!]
\centering
\caption{Lab features of breast cancer patient encounters and corresponding carry forward missing rates.}
\label{tab:labs}
\begin{tabular}{ c c }   
Lab feature & Missing rate \\  
\begin{tabular}[t]{ l} 
\hline
Albumin in blood \\
Alkaline phosphatase in blood \\
Alanine aminotransferase in blood \\
Aspartate aminotransferase in blood \\
Basophils in blood \\
Blood urea nitrogen \\
Calcium level \\
Carbon dioxide level\\
Chloride level \\
Creatinine level in blood \\
Eosinophils in blood \\
Estrogen receptor \\
Glucose level \\
Hematocrit level \\
Hemoglobin level \\
Human epidermal growth factor receptor 2 protein \\
Ki-67 protein \\
Lymphocytes in blood \\
Mean corpuscular hemoglobin concentration \\
Mean corpuscular volume \\
Monocytes in blood \\
Mean platelet volume \\
Neutrophils in blood \\
Pathological node of TNM stage \\
Pathological tumor of TNM stage\\
Platelets in blood\\
P53 gene mutation \\
Potassium level \\
Progesterone receptor \\
Red blood cell count \\
Red cell distribution width \\
Sodium level \\
Tumor grade \\
Tumor size\\
Bilirubin in blood \\
Protein level \\
White blood cell count
\end{tabular} & 
\begin{tabular}[t]{ l} 
\hline
11.0\% \\
12.9\% \\
12.6\% \\
12.6\% \\
13.3\% \\
10.1\% \\
12.4\% \\
12.7\% \\
12.4\% \\
12.4\% \\
12.0\% \\
0\% \\
9.2\% \\
9.8\% \\
9.8\% \\
0\% \\
0\% \\
13.2\% \\
9.7\% \\
9.7\% \\
13.3\% \\
13.1\% \\
13.3\% \\
73.3\% \\
92.9\% \\
9.9\% \\
0\% \\
12.4\% \\
0\% \\
9.8\% \\
9.7\%  \\
12.4 \% \\
61.2\% \\
60.2\% \\
12.9\% \\
12.9\% \\
9.8\%
\end{tabular} \\
\end{tabular}
\end{table}

After physical characteristic averaging, the lab results are the only remaining features with missing values. The initial missing rates for lab results, resulting from a naive encounter-based merge between cocktails and lab results, are as great as 94.9\%. While this is very large, it is consistent with our understanding of patient encounters. Physicians often do not re-order lab tests if there is no reason to expect a change in results. Accordingly, our first-pass data imputation is performing carry forward on all encounters (even those where a cancer drug was not prescribed). After this procedure, nearly all of the missing rates for cocktail-prescribed encounters' lab results fall drastically. We list these missing rates for each lab in Table \ref{tab:labs}. As the final step, we use MICE \cite{mice} to impute all outstanding missing lab results. We run five MICE trials and average the results to create the final, holistic dataset.

To conduct the open-set recognition experiments, we must regard a subset of the breast cancer drug cocktails as the ``novel'' class. Novel drug development signifies an obvious chronology (hence the FDA approval years in Table \ref{tab:cocktails}) and so we designate  the more recent cocktails as ``novel.'' For the purposes of considering well-balanced ``known'' and ``novel'' class splits (in terms of both the number of classes and samples), we conduct two separate experiments. In the first experiment, we designate cocktails 1, 2, 3, and 4 as ``known'' and cocktails 5, 6, and 7 as ``novel'' (cocktail numbers in Table \ref{tab:cocktails}). The second experiment has cocktails 1, 2, and 3 as ``known'' and cocktails 4, 5, 6, and 7 as ``novel.'' More details are given in the respective experimental results subsections. 

Finally for a model-ready dataset, the training, validation, and testing sets are created as follows. The training set is composed of 2/3 of each ``known'' cocktail's samples. The validation set is composed of 1/6 of each of the ``known'' cocktail's samples. These two sets have no novel samples. Finally, the test set is composed of 1/6 of each ``known'' cocktail's samples and a random subset of size 1/6 of each ``novel'' cocktail's samples. In this way, the class balances of each split reflects the population. We create 100 test sets by sampling without replacement within each ``novel'' cocktail. (There are no repeated ``novel'' samples within a test set, but there are across test sets.) Accordingly, we can present test evaluation minimum-to-maximum intervals.

\subsection{Model training and evaluation}
For neural network inputs, the numerical features are normalized and the categorical features are one-hot encoded. We then minimize the loss over the training set (using Adam optimizer with learning rate 0.001) until the objective, evaluated on the known validation set, plateaus or begins to increase. For the numerical features, a Gaussian models the reconstruction. The latent space dimension of $z$ equals 10 for both GMVAE and the ii-loss benchmark. A table of network architectures for GMVAE is presented in Table \ref{tab:GMVAE-networks}. The $\theta$ network is the mirrored $\phi_z$ network. For GMVAE, sigmoid activations follow each hidden layer. The $\phi_z$ network is pretrained on the known classes and the respective weights are then frozen. The $z$ network for the ii-loss benchmark has the same fully connected layers as GMVAE's $\phi_z$ network except for ReLU activations, dropout, and batch normalization layers. We follow their implementation and the details can be found in \cite{inter-intra}. 

 \begin{table}[h!]
\centering
\caption{Network architectures for GMVAE.}
\label{tab:GMVAE-networks}
\begin{tabular}{ ccc }   
$\phi_z$ & $\phi_w$  & $\beta$ \\  
\begin{tabular}[t]{ c } 
\hline
Input: $x$ \\
FC-100 \\
FC-50 \\
FC-20 ($2\times \text{dim}(z)$) \\
\end{tabular} &  
\begin{tabular}[t]{ c} 
\hline
Input: $x,y$ \\
Concatenate with $y$ \\
FC-20 ($2\times \text{dim}(w)$) \\
\end{tabular} & 
\begin{tabular}[t]{ c} 
\hline
Input: $w$ \\
FC-20 \\
FC-20 \\
FC-80 $\left( 2\times C\times \text{dim}(z) \right)$ \\ 
\end{tabular} \\
\end{tabular}
\end{table}

\section{Results} \label{section:results}
From experimental results given next, we clearly demonstrate that GMVAE outperforms the state-of-the-art, deep open-set classifier based on the ii-loss and outlier score, both in terms of accuracy and robustness to an increasing number of novel cocktails (and samples). We attribute this to two primary reasons. First, the GMVAE model also considers reconstruction, which captures additional data structure information, as well as classifier information. Second, GMVAE is more deliberate in algorithmically selecting an ``uncertainty'' threshold $\tau$ based on the known validation set. Indeed, the outlier score in \cite{inter-intra} does not even necessitate a validation set. For this threshold selection as well as model comparison, we utilize the macro-averaged F1 score as our accuracy measure to account for class imbalance.

\subsection{Four existing and three ``novel'' cocktails}
For this first experiment, we divide the cocktails according to Table \ref{tab:known-unknown-cocktails-1}. While ``known'' and ``novel'' splits have a similar number of cocktails, here we are considering a scenario in which there are more ``known'' samples. There are approximately twice as many sample in the ``known'' cocktails as ``novel'' cocktails.

\begin{table}[h!]
\centering
\caption{Breast cancer drug cocktails split into four ``known'' classes and one ``novel'' class according to FDA approval year.}
\label{tab:known-unknown-cocktails-1}
\begin{tabular}{ cccc }   
``Knowns''/ & Cocktail   & Number   & FDA \\  
``novel'' &  numbers   & of  & approval  \\
split & & encounters & year \\
\begin{tabular}[t]{ c } 
\hline
``Knowns'' \\
 \\
 \\
 \\
  \\
``Novel'' \\
 \\

\end{tabular} &  
\begin{tabular}[t]{ l} 
\hline
1 \\
2 \\
3\\
4 \\
 \\
5 \\
6 \\
7 
\end{tabular} & 
\begin{tabular}[t]{ c} 
\hline
2,714 \\
1,868  \\
3,553 \\
5,454  \\
 \\
1,977 \\ 
2,416 \\ 
2,616 
\end{tabular} & 
\begin{tabular}[t]{ c} 
\hline
1991 \\
1991 \\
1992 \\
1998  \\
 \\
2002 \\ 
2003 \\ 
2010 
\end{tabular} \\
\end{tabular}
\end{table}

To contextualize this particular split, we can imagine we are in the year 2000. Can we identify if a patient encounter should receive one of the four ``known'' cocktails (and which one) or should a ``novel'' cocktail be prescribed? The ``novel'' class (composed of three cocktails) is an indication these encounters are opportune for a new, original drug treatment. In reality, we must acknowledge that our study's experimental design is only a proxy to this scenario. We did not enforce samples to this timeline and so there may be ``known'' cocktail patient encounters after say 2003, who actually could have been prescribed one of the ``novel'' cocktails. Such is an inherent limitation of retrospective data used for simulation. Given our already small dataset, we lack the samples prior to 2000 to implement this true timeline. However, our study still captures the relevant, timeless task of classifying which patient encounters should be prescribed an existing versus novel cocktail.

\begin{figure}[h!]
\centerline{
  \includegraphics[width=0.4\linewidth]{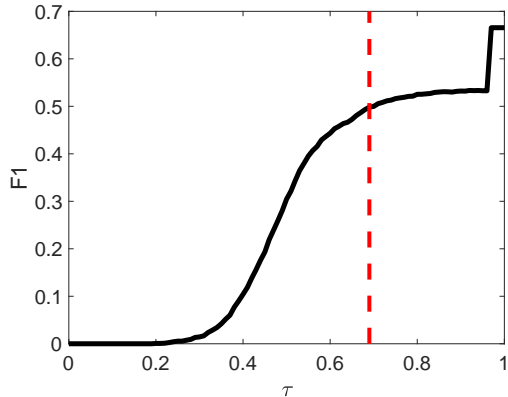} 
}
\caption{Known cocktails validation F1 scores versus GMVAE's ``uncertainty'' threshold $\tau$ and picked threshold $\tau^* = 0.69$ (dashed red line).}
\label{fig:validation-f1s_taus-1}
\end{figure}

As per the authors of \cite{inter-intra}, the outlier score's threshold $\tau$ corresponds with a 1\% contamination rate. We now detail our procedure for selecting the ``uncertainty'' threshold. Plotted in Figure \ref{fig:validation-f1s_taus-1} are the known validation F1 scores versus $\tau$ for GMVAE's $U$ quantity. Work in \cite{my-gmvae} deduces (and empirically observes) that a consistently good threshold $\tau$ to pick for GMVAE's ``uncertainty'' is the saturation or plateau point of the known validation F1 curve. This is plotted with the red dashed line in Figure \ref{fig:validation-f1s_taus-1}. Intuitively, this can be thought of as increasing the decision boundary around each class's centroid until diminishing classification accuracy returns. Further increasing $\tau$ is tantamount to overfitting the known validation samples and risks under-recognizing ``novel'' samples. Mathematically we define this saturation point in the following way. Letting 
\begin{align}
\widetilde{\tau} = \min \left\{ \tau: \text{F1}'(\tau)\geq \epsilon_1 \right\}~,
\end{align}
then the selected threshold is set to the saturation point
\begin{align}
\tau^* = \min \left\{ \tau: \tau > \widetilde{\tau} \quad \text{and} \quad \text{F1}'(\tau)\leq \epsilon_2 \right\}~.
\end{align}
For this experiment, we use $\epsilon_1=1$ and $\epsilon_2=0.25$ and approximate $\text{F1}'(\tau)$ by using the simple forward difference scheme.

\begin{figure}[h!]
\centerline{
  \includegraphics[width=0.4\linewidth]{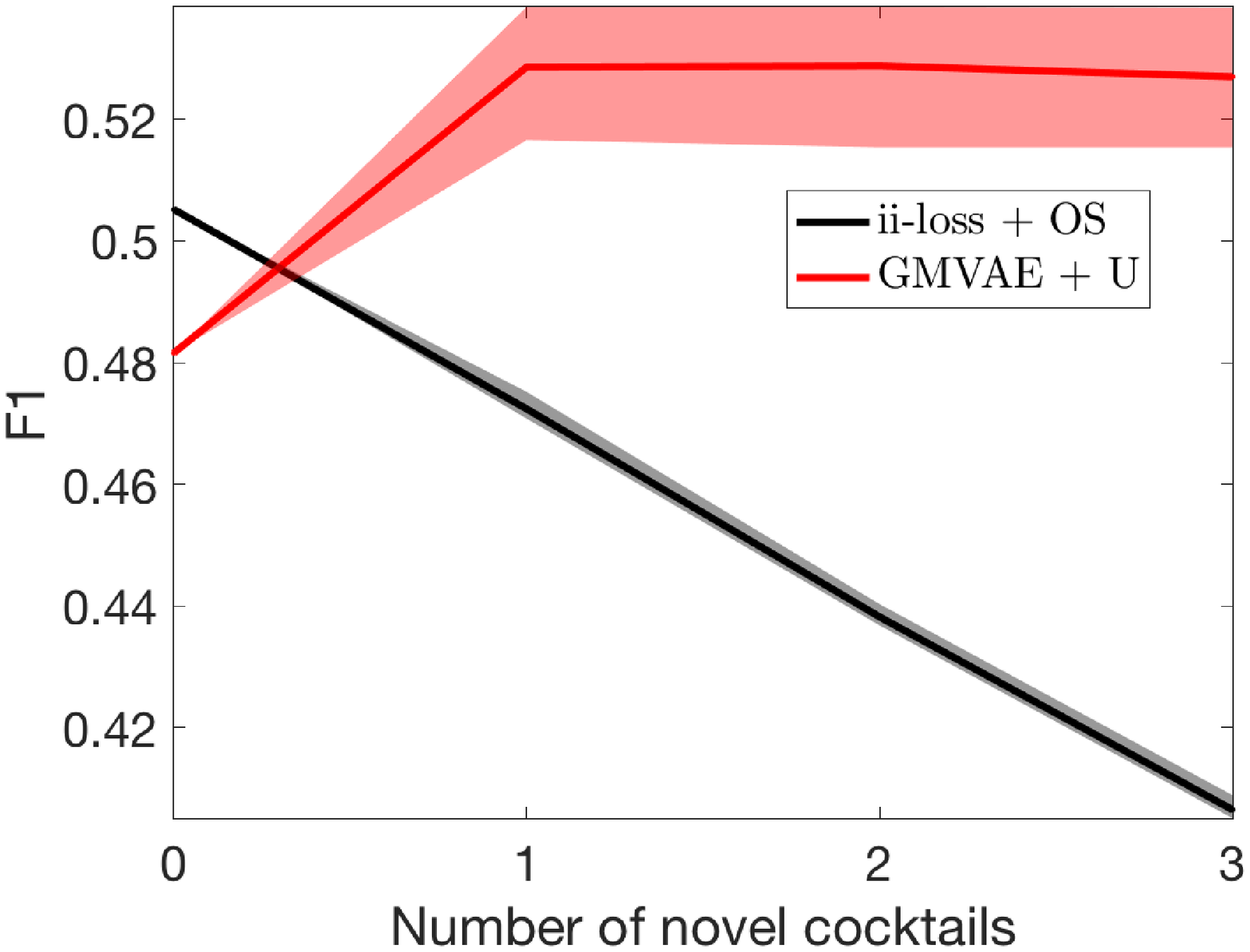} 
  \includegraphics[width=0.4\linewidth]{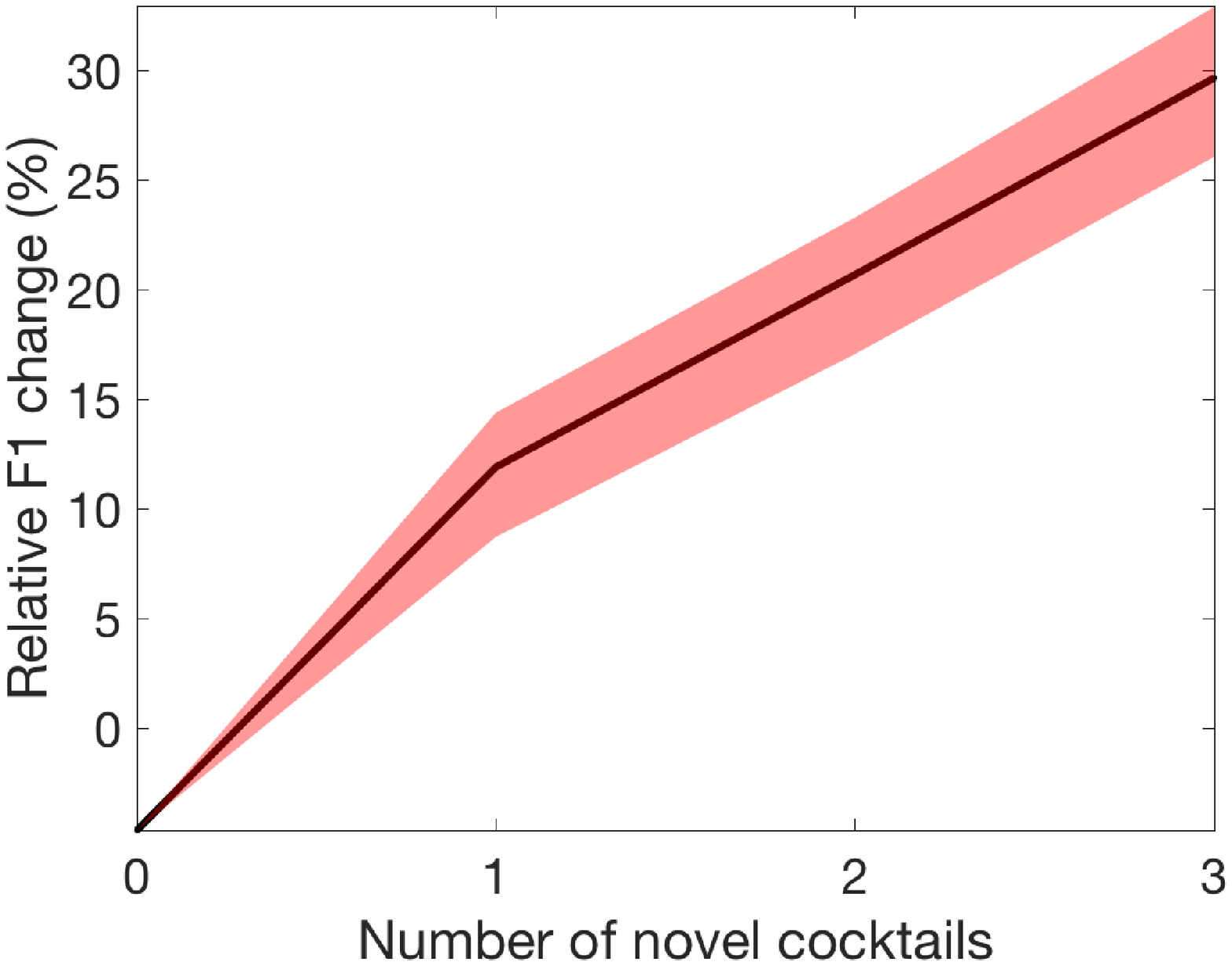} 
}
\caption{(Left) Open-set test F1 score intervals versus number of ``novel'' cocktails. Shaded bands show maximums and minimums for the respective points. (Right) GMVAE + U's relative changes in F1 scores compared to ii-loss + OS.}
\label{fig:compare-f1s-1}
\end{figure}

With the selected threshold $\tau^*$ for ``uncertainty,'' we proceed to the testing phase with ``novel'' cocktails. To study robustness to increasing ``novel'' samples, as well as accuracy, we incrementally increase the number of ``novel'' cocktails (according to the order in Table \ref{tab:known-unknown-cocktails-1}) and measure F1 scores. These test F1 scores versus the number of ``novel'' cocktails are plotted in Figure \ref{fig:compare-f1s-1}. As previously discussed, GMVAE is not as accurate in the closed-set regime with no ``novel'' samples because the ii-loss more directly optimizes known-class discrimination. However, GVMAE and ``uncertainty'' (GMVAE + U) quickly outperform ii-loss and outlier score (ii-loss + OS) with the introduction of ``novel'' cocktails. In addition, we clearly see that the GMVAE's method remains more robust to an increasing number of ``novel'' cocktails and samples while the benchmark outlier score's accuracy continuously diminishes. This observation is magnified in the right panel of Figure \ref{fig:compare-f1s-1}. Averaging over the number of ``novel'' cocktails, GMVAE leads to a 14.4\% increase in the F1 score. Again, we attribute this increased accuracy and robustness to GMVAE's reconstruction learning and ``uncertainty's'' penalization of interior latent representations relative to the known clusters. From \cite{my-gmvae}, we surmise the latter's effect is substantial for more homogeneous, difficult-to-discriminate samples. This is certainly true of healthcare data.

While F1 scores paint a broad picture of classification accuracy, the confusion matrices in Table \ref{tab:confusion-matrices-1} more closely inspect prediction ability on an individual cocktail basis. We clearly see that ii-loss + OS is more accurate in the closed-set regime. In particular, cocktails 3  and 4 are classified very accurately. However, this comes at the expense of severely under-recognizing ``novel'' cocktails. The ii-loss + OS model rarely predicts ``novel'' leading to a dramatic decrease in overall open-set classification accuracy. No doubt this is due, in part, to ii-loss solely optimizing known cocktail discrimination with no regard for capturing underlying feature information. Conversely, GMVAE is less accurate in the closed-set regime but more readily recognizes ``novel'' cocktails. The main point these confusion matrices demonstrate is that there is a tradeoff between accurately classifying the known classes and robustly identifying novel or unknown classes. This is evidenced by GMVAE's over-classifying the known cocktails as ``novel.''

\begin{table}[h!]
\centering
\caption{Single test set confusion matrices with true cocktail class as rows and predicted cocktail as columns for (left) ii-loss + OS and (right) GMVAE. Double horizontal line divides known and novel cocktails.}
\label{tab:confusion-matrices-1}
\begin{tabular}{c|c|c|c|c|c|c|}
\cline{3-7}
\multicolumn{2}{c|}{\textbf{Cocktails}}&\textbf{1}&\textbf{2}&\textbf{3} & \textbf{4} & \textbf{Novel}\\
\cline{2-7}
& \textbf{1} & 213 & 117 & 54 & 64 & 5 \\
\cline{2-7}
& \textbf{2} & 114 & 115 & 37 & 31 & 15 \\
\cline{2-7}
& \textbf{3} & 16 & 14 & 468 & 95 & 0 \\
\cline{2-7}
& \textbf{4} &21 & 0 & 92 & 796 & 0 \\
\hhline{~|=|=|=|=|=|=|}
& \textbf{5} & 40 & 15 & 156 & 118 & 0 \\
\cline{2-7}
& \textbf{6} & 45 & 8 & 221 & 128 & 0 \\
\cline{2-7}
& \textbf{7} & 43 & 41& 212	 & 140& 0 \\
\cline{2-7}
\end{tabular} \qquad \begin{tabular}{c|c|c|c|c|c|c|}
\cline{3-7}
\multicolumn{2}{c|}{\textbf{Cocktails}}&\textbf{1}&\textbf{2}&\textbf{3} & \textbf{4} & \textbf{Novel}\\
\cline{2-7}
& \textbf{1} & 163 & 117 & 33 & 20 & 120 \\
\cline{2-7}
& \textbf{2} & 101 & 110& 21 & 10 & 70 \\
\cline{2-7}
& \textbf{3} & 0 & 0 & 393 & 60	 & 140 \\
\cline{2-7}
& \textbf{4} & 3 & 2 & 49 & 704 & 151 \\
\hhline{~|=|=|=|=|=|=|}
& \textbf{5} & 3 & 2 & 98 & 78 & 148 \\
\cline{2-7}
& \textbf{6} & 7 & 2 &  131& 86 & 176 \\
\cline{2-7}
& \textbf{7} & 3 & 2& 97 & 98 & 236 \\
\cline{2-7}
\end{tabular}
\end{table}

\begin{figure}[h!]
\centerline{
  \includegraphics[width=0.4\linewidth]{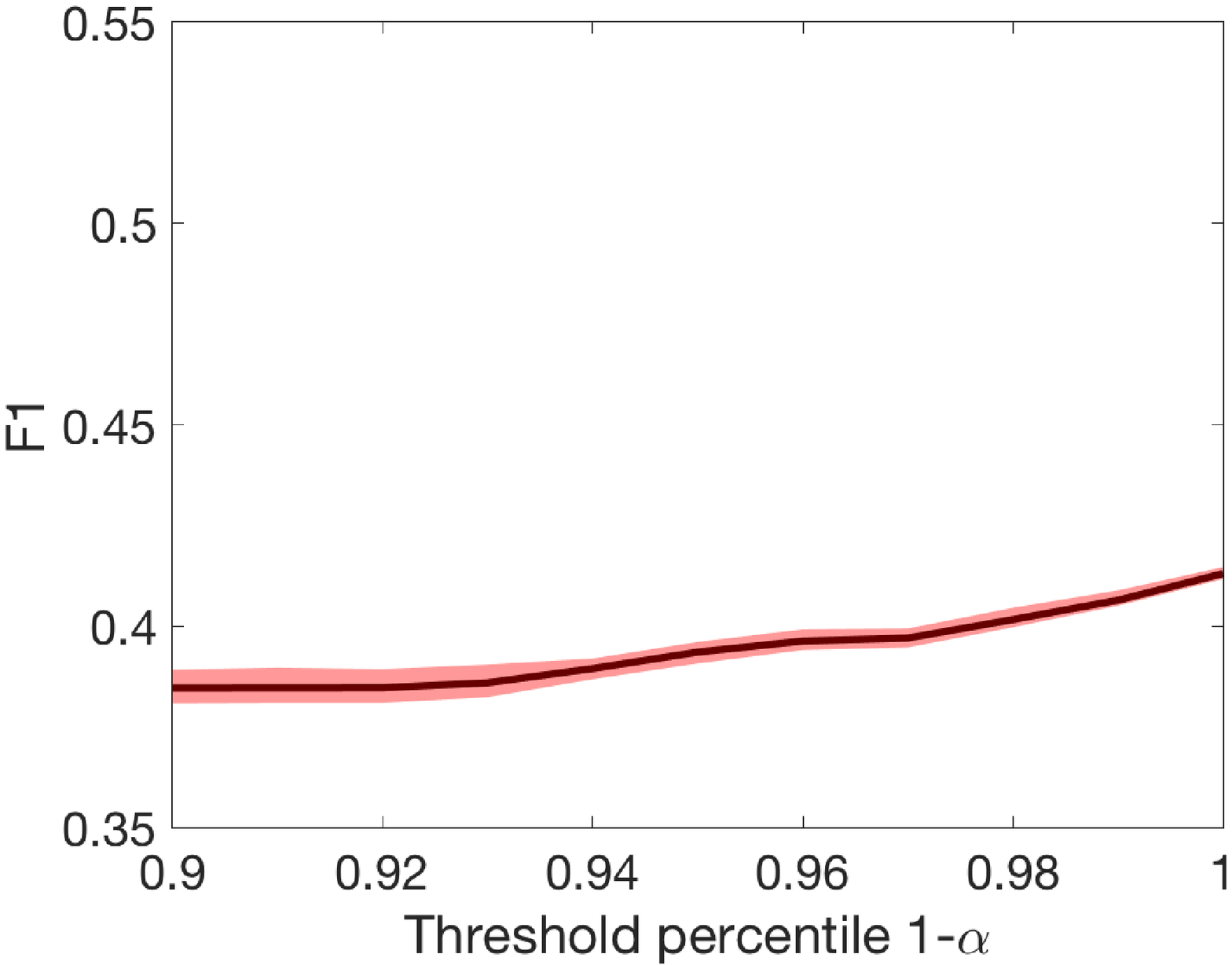} 
    \includegraphics[width=0.4\linewidth]{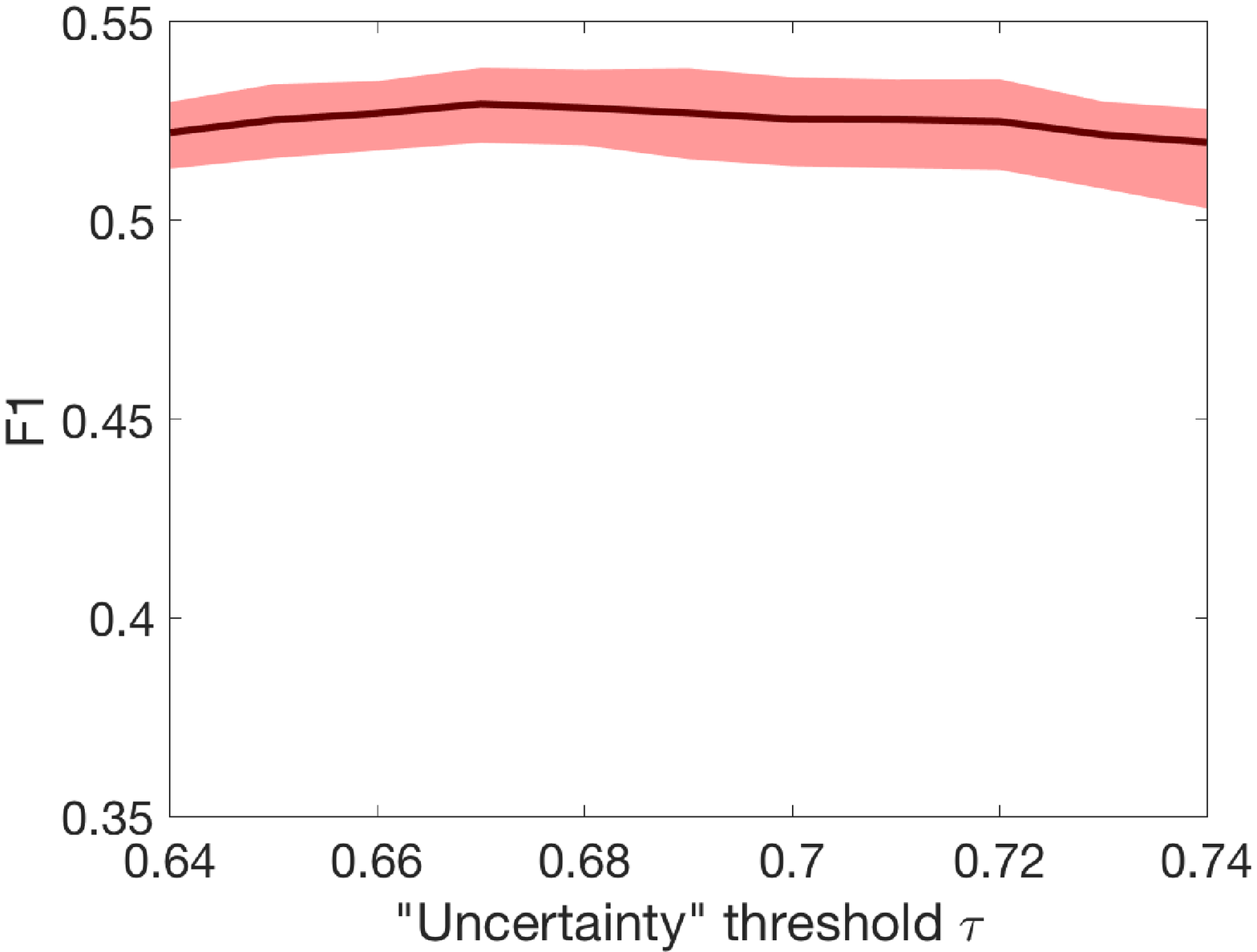} 
}
\caption{Open-set test F1 score intervals, calculated using all ``novel'' cocktails, versus nearby neighborhood of (left) ii-loss + OS's $\alpha$ and (right) GMVAE's $\tau^*$. Red shaded bands show maximums and minimums for the respective points.}
\label{fig:tau-sweep-1}
\end{figure}

Finally, we wish to further address the threshold selection. To alleviate concerns that open-set accuracies are more sensitive to threshold selection, we plot the test F1 scores versus thresholds in neighborhoods of $\alpha$ and $\tau^*$ for ii-loss + OS and GMVAE, respectively in Figure \ref{fig:tau-sweep-1}. We consider a 10 percentage point neighborhood so that $\alpha \in [0,0.1]$ and $\tau^* \in [0.64, 0.74]$. We clearly see that GMVAE's F1 scores are more constant with respect to $\tau^*$ than ii-loss + OS's F1 versus $\alpha$. This translates to GMVAE being more robust to threshold selection (error) and, relatedly, its underlying embedding having a stronger ability to distinguish ``novel'' cocktails.

\subsection{Three existing and four ``novel'' cocktails}
For this second experiment, we divide the cocktails according to Table \ref{tab:known-unknown-cocktails-2}. While ``known'' and ``novel'' splits again have a similar number of cocktails as in the previous experiment, here we increase the number of ``novel'' cocktails by one and consider the different scenario in which there are more ``novel'' samples. There are approximately 1.5 times as many samples in the ``novel'' cocktails as ``known'' cocktails.

\begin{table}[h!]
\centering
\caption{Breast cancer drug cocktails split into three ``known'' classes and one ``novel'' class according to FDA approval year.}
\label{tab:known-unknown-cocktails-2}
\begin{tabular}{ cccc }   
``Knowns''/ & Cocktail   & Number   & FDA \\  
``novel'' &  numbers   & of  & approval  \\
split & & encounters & year \\
\begin{tabular}[t]{ c } 
\hline
``Knowns'' \\
 \\
 \\
  \\
``Novel'' \\
 \\
 \\

\end{tabular} &  
\begin{tabular}[t]{ l} 
\hline
1 \\
2 \\
3\\
 \\
4 \\
5 \\
6 \\
7 
\end{tabular} & 
\begin{tabular}[t]{ c} 
\hline
2,714 \\
1,868  \\
3,553 \\
 \\
 5,454  \\
1,977 \\ 
2,416 \\ 
2,616 
\end{tabular} & 
\begin{tabular}[t]{ c} 
\hline
1991 \\
1991 \\
1992 \\
 \\
 1998  \\
2002 \\ 
2003 \\ 
2010 
\end{tabular} \\
\end{tabular}
\end{table}

\begin{figure}[h!]
\centerline{
  \includegraphics[width=0.4\linewidth]{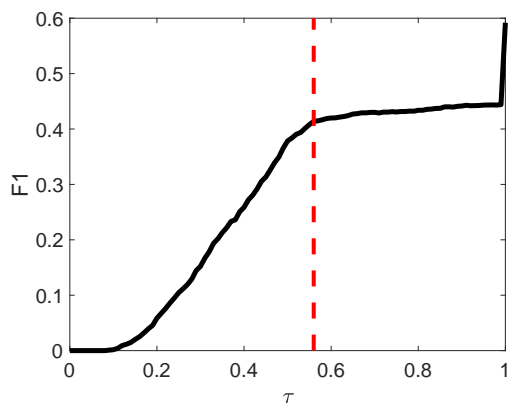} 
}
\caption{Second experiment's known cocktails validation F1 scores versus GMVAE's ``uncertainty'' threshold $\tau$ and picked threshold $\tau^* = 0.56$ (dashed red line).}
\label{fig:validation-f1s_taus-2}
\end{figure}

\begin{figure}[h!]
\centerline{
  \includegraphics[width=0.4\linewidth]{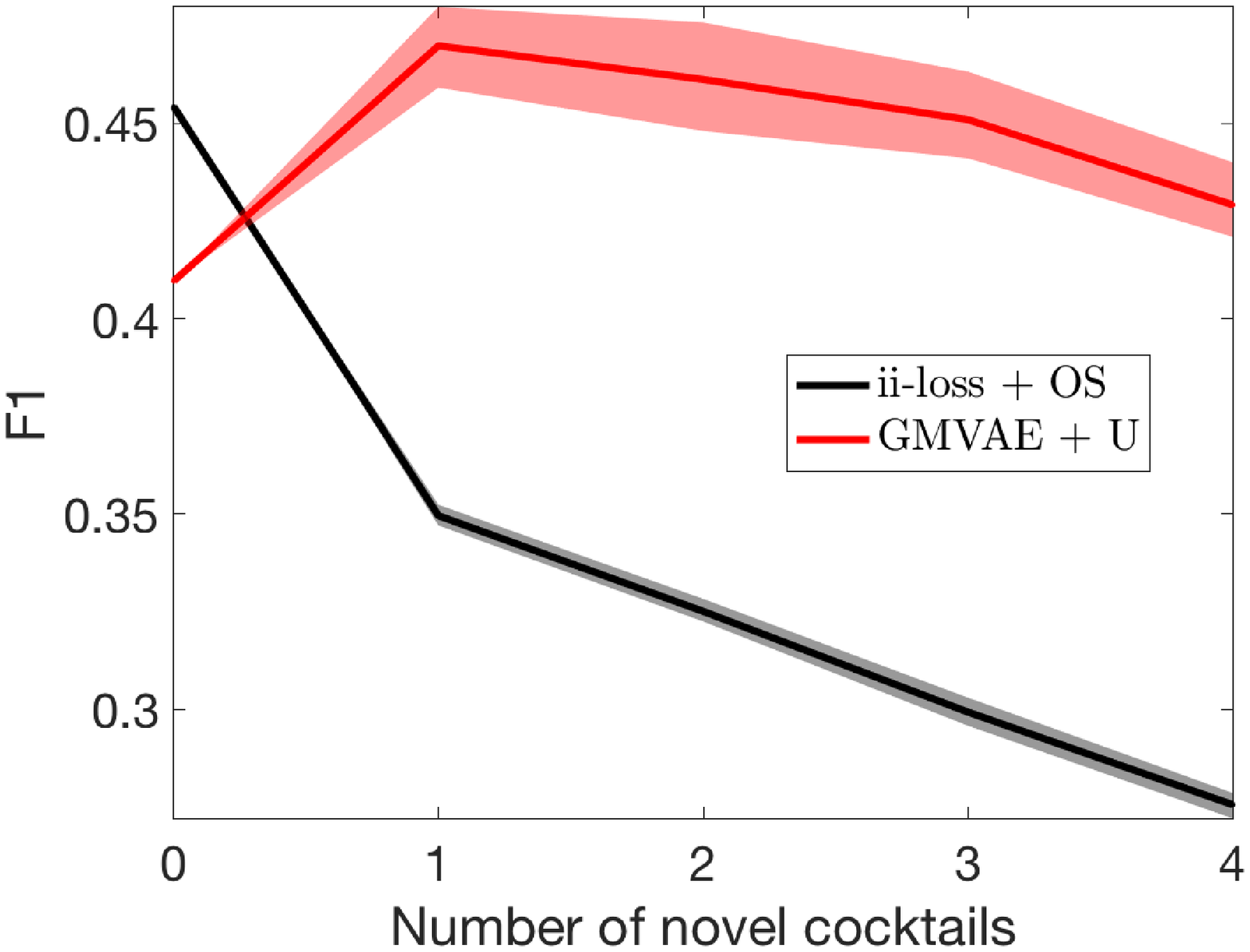} 
  \includegraphics[width=0.4\linewidth]{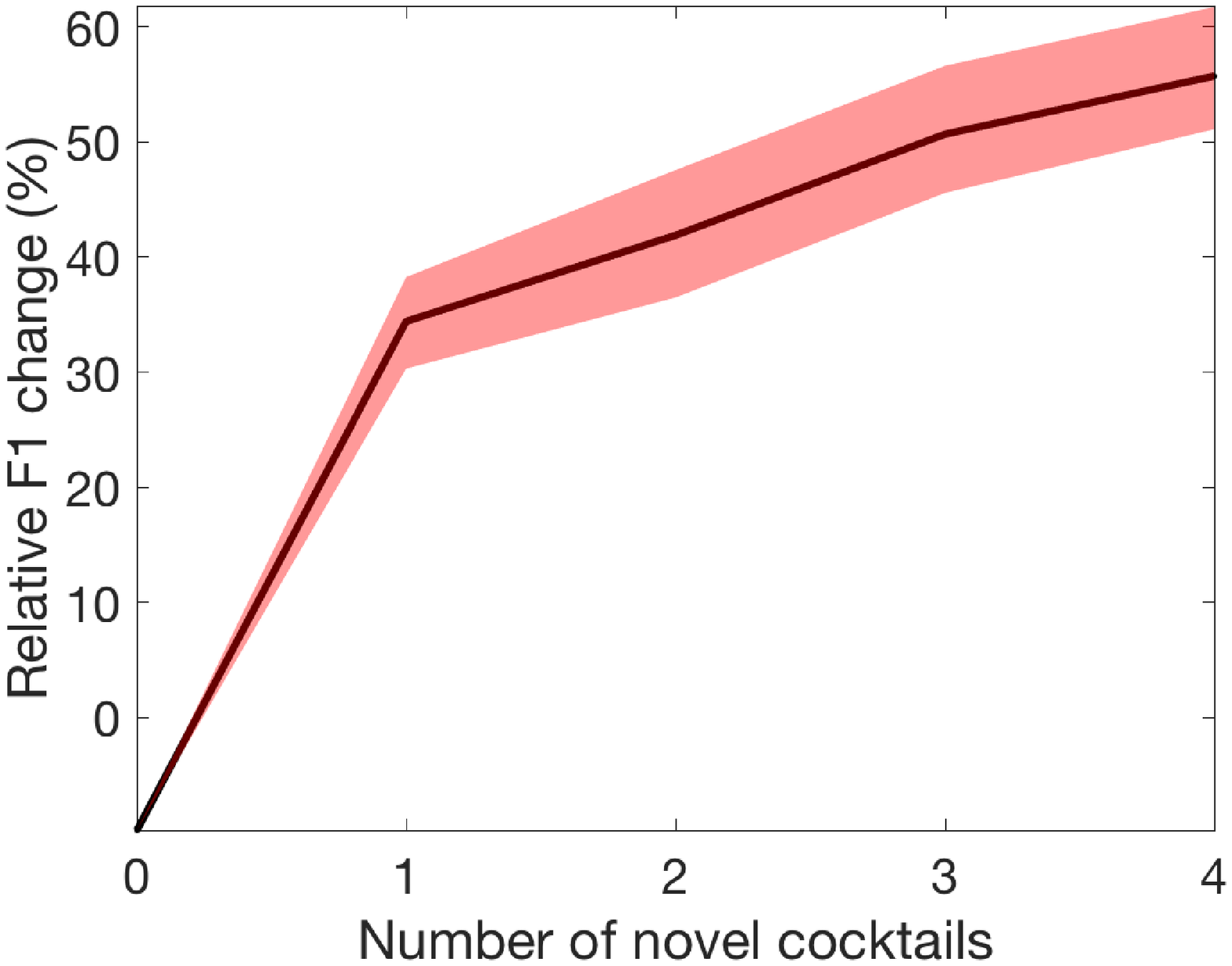} 
}
\caption{(Left) Second experiment's open-set test F1 score intervals versus number of ``novel'' cocktails. Shaded bands show maximums and minimums for the respective points. (Right) GMVAE + U's relative changes in F1 scores compared to ii-loss + OS.}
\label{fig:compare-f1s-2}
\end{figure}

\begin{figure}[h!]
\centerline{
  \includegraphics[width=0.4\linewidth]{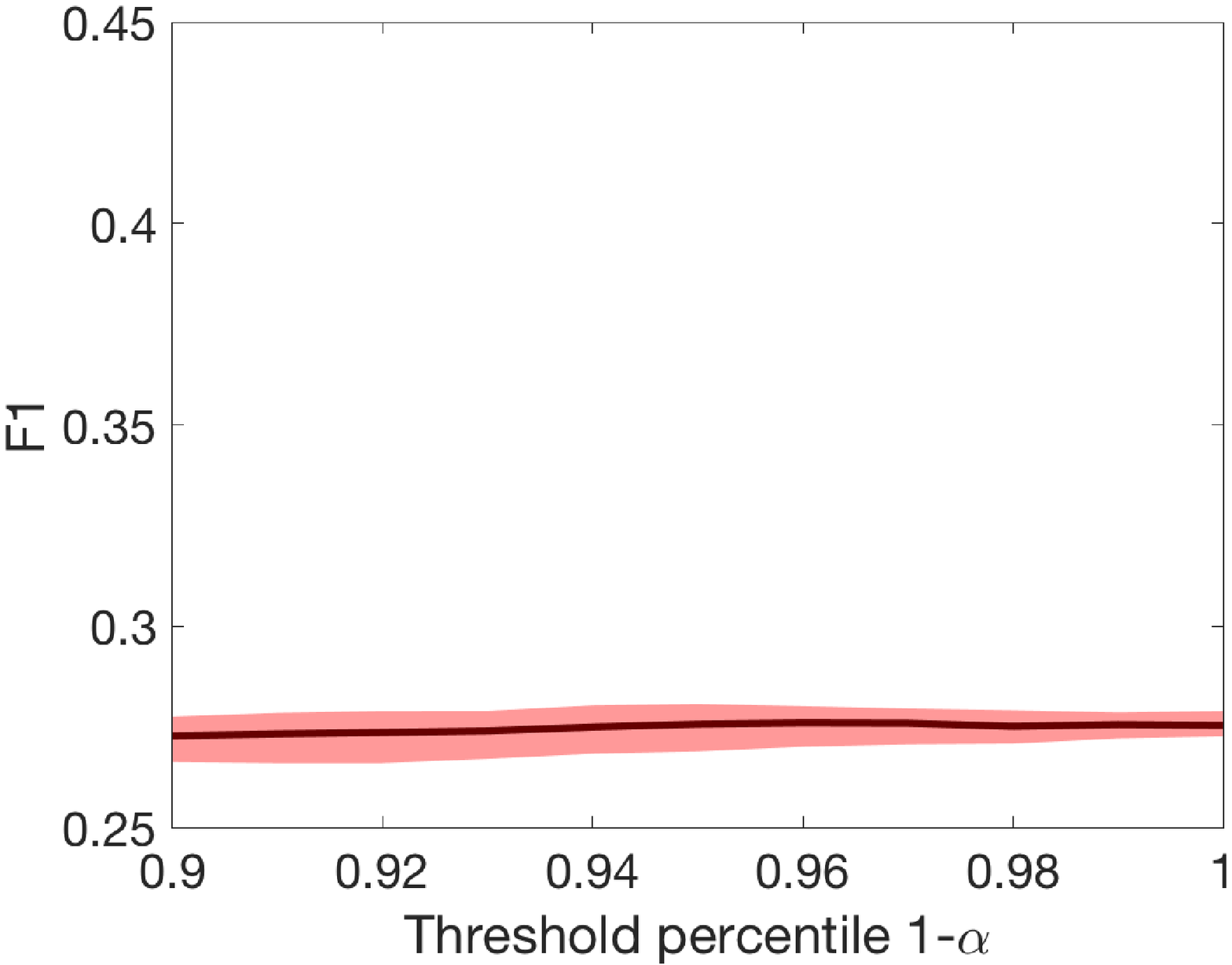} 
    \includegraphics[width=0.4\linewidth]{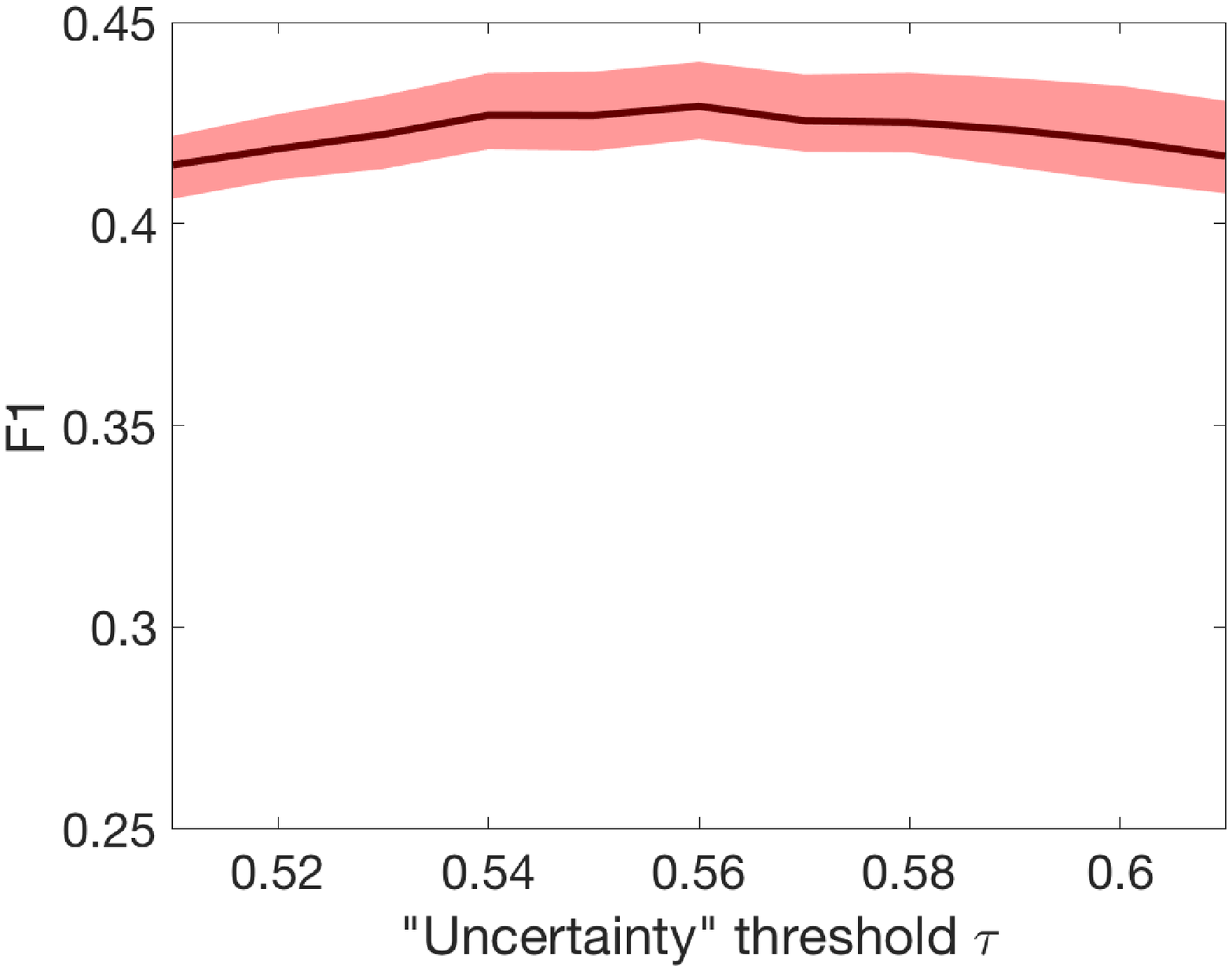} 
}
\caption{Second experiment's open-set test F1 score intervals, calculated using all ``novel'' cocktails, versus nearby neighborhood of (left) ii-loss + OS's $\alpha$ and (right) GMVAE's $\tau^*$. Red shaded bands show maximums and minimums for the respective points.}
\label{fig:tau-sweep-2}
\end{figure}

The goal of this second experiment is to reiterate GMVAE's success  while varying the number of ``known'' cocktails and having a qualitatively different ``known''-to-``novel'' samples ratio. Parallel Figures \ref{fig:validation-f1s_taus-2}-\ref{fig:tau-sweep-2} and Table \ref{tab:confusion-matrices-2} from the first experiment show just this. Figure \ref{fig:validation-f1s_taus-2} again plots the known validation F1 scores versus $\tau$ for GMVAE's $U$ quantity with the saturation point and corresponding picked threshold $\tau^*$ plotted in red. 

Again, we incrementally increase the number of ``novel'' cocktails (according to the order in Table \ref{tab:known-unknown-cocktails-2}) and plot the F1 scores in Figure \ref{fig:compare-f1s-2}. The behavior is qualitatively the same as the first experiment. The ii-loss + OS is more accurate with just the ``known'' cocktails as it directly optimizes class separation. However, GMVAE+U yields much higher open-set classification accuracies for increasing ``novel'' cocktails and samples. In fact, averaging over the number of ``novel'' cocktails in Figure \ref{fig:compare-f1s-2}'s right panel, GMVAE+U leads to an average F1 increase of 34.6\%. We stress that this increased open-set recognition is from GMVAE's reconstruction learning and ``uncertainty'' leading to better discernment of ``novel'' cocktails. This is made clear in the confusion matrices in Table \ref{tab:confusion-matrices-2}. The ii-loss + OS's $z$ embedding only captures class information and therefore it is difficult to tease ``novel'' cocktail information from.

\begin{table}[h!]
\centering
\caption{Second experiment's single test set confusion matrices with true cocktail class as rows and predicted cocktail as columns for (left) ii-loss + OS and (right) GMVAE. Double horizontal line divides known and novel cocktails.}
\label{tab:confusion-matrices-2}
\begin{tabular}{c|c|c|c|c|c|}
\cline{3-6}
\multicolumn{2}{c|}{\textbf{Cocktails}}&\textbf{1}&\textbf{2}&\textbf{3}  & \textbf{Novel}\\
\cline{2-6}
& \textbf{1} & 288 & 125 & 33 & 7  \\
\cline{2-6}
& \textbf{2} & 133 & 122 & 52 & 5  \\
\cline{2-6}
& \textbf{3} & 17 & 107 & 456 & 13  \\
\hhline{~|=|=|=|=|=|}
& \textbf{4} &187 & 427 & 292 & 3  \\
\cline{2-6}
& \textbf{5} & 35 & 133 & 153 & 8  \\
\cline{2-6}
& \textbf{6} & 41 & 156 & 202 & 3 \\
\cline{2-6}
& \textbf{7} & 79 & 185 & 167	 & 5 \\
\cline{2-6}
\end{tabular} \qquad \begin{tabular}{c|c|c|c|c|c|}
\cline{3-6}
\multicolumn{2}{c|}{\textbf{Cocktails}}&\textbf{1}&\textbf{2}&\textbf{3}  & \textbf{Novel}\\
\cline{2-6}
& \textbf{1} & 231 & 55 & 48 & 119  \\
\cline{2-6}
& \textbf{2} & 129 & 57 & 42& 84  \\
\cline{2-6}
& \textbf{3} & 2 & 1 & 459 & 131  \\
\hhline{~|=|=|=|=|=|}
& \textbf{4} &37 & 24 & 446 & 402  \\
\cline{2-6}
& \textbf{5} & 8 & 7 & 172 & 142  \\
\cline{2-6}
& \textbf{6} & 3 & 1 & 191 & 207 \\
\cline{2-6}
& \textbf{7} & 9 & 3 & 302 & 122 \\
\cline{2-6}
\end{tabular}
\end{table}

To again show robustness to threshold selection, we plot the test F1 scores versus thresholds in neighborhoods of $\alpha$ and $\tau^*$ for ii-loss + OS and GMVAE respectively in Figure \ref{fig:tau-sweep-2}. We consider a 10 percentage point neighborhood so that $\alpha \in [0,0.1]$ and $\tau^* \in [0.51, 0.61]$. Here we clearly see that both the ii-loss + OS's and GMVAE's F1 scores are relatively constant with respect to $\alpha$ and $\tau^*$. However, the difference in F1 is stark with GMVAE dominating.

\section{Limitations and future work} \label{section:limitations-future_work}
While the experimental results highlight GMVAE's capability, we do wish to stress again the importance of methodically selecting a single threshold for rejecting ``novel'' samples in testing. Previous open-set experiments \cite{oct_images, diagnoses, inter-intra} escape this consideration by comparing high-level metrics like AUC. While this may give an indication of the overall behavior, it does little to inform actual model usage in a practical setting. GMVAE's validation F1 curve saturation procedure begins to address this decision boundary optimization without ``novel'' samples. However, it is still an ad-hoc heuristic worthy of further development. A satisfying solution to this subproblem is critical for real applications to a current patient's treatment plan.

\begin{figure}[h!]
\centerline{
  \includegraphics[width=0.5\linewidth]{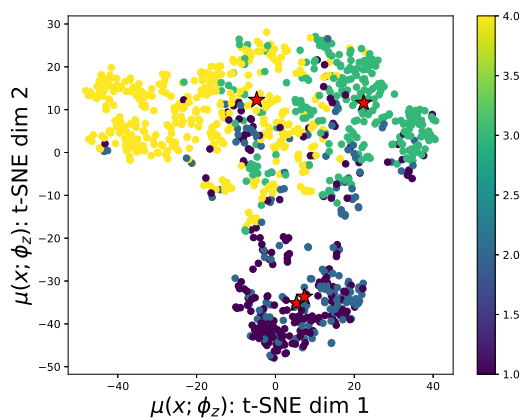} 
}
\caption{t-SNE plot of $\mu(x; \phi_z )$ from the first experiment's GMVAE training latent representations for the known cocktails. Colors distinguish cocktails and red stars are the class centroids.}
\label{fig:train-tsne}
\end{figure}

Additional discussions of our current application of open-set recognition to drug treatment predictions are more subtle. From the experiments above, while we achieve more accurate results, an F1 score below 0.5 has room for improvement. Prior work and our own experience suggest there exists a tradeoff between closed-set classification and open-set recognition. In other words, it is natural to expect a compromise between accurately classifying the ``known'' classes and robustly identifying ``novel'' or ``unknown'' classes. Herein lies this issue, as it is generally more difficult to discriminate real healthcare data (as opposed to academic image datasets), we start at a severe disadvantage with separating the known cocktails. We visualize this with a t-SNE \cite{t-sne} plot of GMVAE's latent embedding from the first experiment in Figure \ref{fig:train-tsne}. It indicates there is high degree of feature ``overlap'' among the cocktails and thus it is difficult to distinguish patient encounters. This is likely expected from only utilizing the phenotypic features available in the dataset, but perhaps also from the nature of patients potentially benefiting from multiple cocktails. In that respect, open-set recognition within a multi-label setting may be a natural extension we may pursue in future work.

Lastly, perhaps drug treatment predictions, at least in our cancer context, necessitate a longitudinal study. One would like to track multiple encounters for a given patient and be able to make breast cancer drug treatment recommendations along the way. With this framework, the efficacy feedback information from past treatments could potentially provide significant guidance. While this would require a complete overhaul from data collection to modifying the model to accept a variable-length time series, it certainly represents an obvious and exciting expansion. Relatedly, this immediately suggests the addition of multi-modal data such as mammogram images and physician notes' text in a recurrent neural network model structure.

\section{Conclusion} \label{section:conclusion}
We formulate breast cancer drug cocktail treatment predictions in terms of open-set recognition, focusing on a methodology conducive to a practical clinical implementation, and accordingly apply the GMVAE and ``uncertainty'' model framework. Together, these achieve more accurate and robust classification results for our patient-encounter healthcare dataset, compared to a state-of-the-art benchmark. In doing so, we also resolve obstacles in prior work concerning open-set recognition applications to healthcare. First, we emphasize formally addressing a methodical selection of a specific threshold for rejecting ``novel'' or ``unknown''  samples as we believe it is more meaningful in deployment to compare and use models with a single testing instance. Second, many other works on this subject take advantage of fabricated or auxiliary data to model ``novel'' or  ``unknown'' samples. We dismiss the implicit assumption that this step is always feasible and instead call for methods like GMVAE which only learn from known, available data. Finally, we spotlight the inherent limitations to solely classification-based models in open-set recognition. Whether it is reconstruction or not, latent representations must encapsulate structural information of the data to be more effective. To be sure, this particular application to healthcare opens interesting avenues of further research to the expanding scope of open-set recognition. Likewise, this study hopefully represents a stride towards these techniques benefiting actual patients' treatments in the future.

\section*{Acknowledgments}
The work of the first author is supported by the Predoctoral Training Program in Biomedical Data Driven Discovery (BD3) at Northwestern University (National Library of Medicine Grant 5T32LM012203). The work of the last author is supported in part by NIH Grant R01LM013337.

\section*{Conflicts of interest statement}
The authors declare there are no conflicts of interest.

\section*{Appendix}

\subsection*{Background for GMVAE and ``uncertainty'' algorithm}
Here in this Appendix section, we briefly overview GMVAE, which extends standard unsupervised VAEs by assuming a Gaussian mixture prior for each class. To accommodate this, the basic VAE architecture is modified with additional latent variables. \cite{my-gmvae} starts with $C$ known classes with each class composed of $K_c$ mixture components where $c = 1, 2, ..., C$. The features $x \in \R^d$ and labels $y\in\R^C$, represented as one-hot vectors, comprise the labeled, known dataset. GMVAE's decoder model $p_{\beta,\theta}(x,v,w,z|y) = p_\theta (x|z) p_\beta(z|w,y,v)p(w) p(v|y)$ conditions on class and factors as
\begin{align}
w &\sim \mathcal{N} (0, I) \\ 
(v | y ) \in \R^{K_c} &\sim \text{Mult} (\pi(y)) \\
(z|w,y,v) &\sim \prod_{c=1}^C \prod_{k=1}^{K_c} \mathcal{N} \left(\mu_{ck} (w; \beta), \text{diag}\left( \sigma^2_{ck} (w; \beta)\right) \right)^{y_c\cdot v_k} \\
(x | z) &\sim \mathcal{B} \left( \mu(z; \theta)\right)
\end{align}
where $\mu_{ck} (\cdot; \beta)$, $\sigma^2_{ck} (\cdot; \beta)$, and $\mu(\cdot; \theta)$ are neural networks parametrized by $\beta$ and $\theta$, respectively. It is common to assume a uniform prior $\pi(y)$. The encoder process is factorized as $q_\phi (v,w,z|x,y)  = p_\beta (v | z,w,y)q_{\phi_w}(w|x,y)q_{\phi_z}(z|x)$ where $\phi = (\phi_x, \phi_w)$. Factors $\phi$ are parametrized with networks that output mean and diagonal covariance for Gaussian posteriors:
\begin{align}
(z|x) & \sim \mathcal{N}  \left( \mu(x; \phi_z), \text{diag} \left( \sigma^2(x; \phi_z) \right) \right)  \\
(w|x,y) & \sim \mathcal{N}  \left( \mu(x,y; \phi_w), \text{diag} \left( \sigma^2(x,y; \phi_w) \right) \right). 
\end{align}
There is a $p_\beta$ factor in the $q_\phi$ factorization because it is derived from the generative factors (see \cite{my-gmvae}). GMVAE's objective is to maximize the log-evidence lower bound (ELBO) given by
\begin{align}
&\mathcal{L} (K)  = \E_{q_\phi (v,w,z|x,y)} \left[ \log \frac{p_{\beta,\theta}(x,v,w,z|y)}{q_\phi (v,w,z|x,y)} \right] \\
&= \E_{q_{\phi_z} (z|x)}\left[ \log p_\theta (x|z) \right] \quad \text{(reconstruction)} \\
&- \mathbb{E}_{q_{\phi_w}(w|x,y) q_{\phi_z}(z|x)  } \left[ \log q_{\phi_z}(z|x)  - \sum_{j=1}^{K_c} p_\beta ( v=j|z, w, y)  \log p_\beta ( z|w, y, v=j)  \right] \quad \text{(latent covering)} \\
&- KL(q_{\phi_w}(w|x,y) || p(w) ) \quad \text{($w$-prior)}\\
&- \mathbb{E}_{q_{\phi_w}(w|x,y) q_{\phi_z}(z|x)} \left[ KL( p_\beta ( v|z, w, y)  || p (v|y) ) \right] \quad  \text{(component $v$-prior)}.
\end{align}
Vector $K=(K_1, K_2, ..., K_C)$ is decided by the user and so the ELBO dependence on $K$ is made explicit. The reconstruction term endeavors to group samples with similar features together in latent space $z$. Simultaneously, the latent covering term attempts to cluster the latent representations $z$ based on classes. The $w$-prior and component $v$-prior terms aim for the respective posteriors and  priors to coincide. This mirrors the standard ELBO with reconstruction and regularization terms.

\bibliographystyle{unsrt}
\bibliography{cancer-open_set}

\end{document}